\documentclass{article}
\PassOptionsToPackage{numbers, compress}{natbib}
 \usepackage[preprint]{neurips_2026}

\usepackage[utf8]{inputenc} 
\usepackage[T1]{fontenc}    
\usepackage{hyperref}       
\usepackage{url}            
\usepackage{booktabs}       
\usepackage{amsfonts}       
\usepackage{nicefrac}       
\usepackage{microtype}      
\usepackage{xcolor}         

\usepackage{amsmath}
\usepackage{natbib}
\usepackage{booktabs}
\usepackage{multirow}
\usepackage{tabularx}

\usepackage{graphicx}
\usepackage{caption}
\usepackage{subcaption} 
\usepackage{hyperref}
\usepackage{cleveref}   

\usepackage{hyperref}
\hypersetup{
    colorlinks=true,
    linkcolor=blue,
    filecolor=magenta,      
    urlcolor=cyan,
    citecolor=black
}

\usepackage{enumitem}

\usepackage{titlesec}
\titlespacing\section{0pt}{0pt plus 0pt minus 1pt}{0pt plus 0pt minus 1pt}
\titlespacing\subsection{0pt}{0pt plus 0pt minus 1pt}{0pt plus 0pt minus 1pt}
\titlespacing\subsubsection{0pt}{0pt plus 0pt minus 1pt}{0pt plus 0pt minus 1pt}

\AtBeginDocument{%
  \addtolength\abovedisplayskip{-0.1\baselineskip}%
  \addtolength\belowdisplayskip{-0.1\baselineskip}%
 \addtolength\abovedisplayshortskip{-0.1\baselineskip}%
 \addtolength\belowdisplayshortskip{-0.1\baselineskip}%
}

\title{Topological Steering}

\author{%
  Benoît Guérand\\
  Department of Mathematics\\
  National University of Singapore (NUS)\\
  2 Science Drive 2, Singapore 117543 \\
  \texttt{benoit.guerand@u.nus.edu} \\
\And
  Tan Minh Nguyen \\
  Department of Mathematics\\
  National University of Singapore (NUS)\\
  2 Science Drive 2, Singapore 117543 \\
  \texttt{tanmn@nus.edu.sg} \\
}

\begin{document}

\maketitle

\begin{abstract}
With the rapid rise of large language models (LLMs), controlling undesirable model behaviors has become increasingly important. Existing behavioral control methods typically intervene directly in activation or feature space, but such approaches can be sensitive to outliers, distributional shifts, noise, and other local perturbations. Motivated by Topological Data Analysis (TDA), which captures global rather than purely local structure, we propose Topological Steering, a new framework for steering LLM behavior through the topological representation of activation spaces. Using persistence diagrams, our method connects activation-based steering with TDA and enables more robust behavioral control. We show that Topological Steering consistently modifies LLM behavior across multiple model families and model sizes.
\end{abstract}

\section{Introduction}\label{sec:Introduction}
Large language models (LLMs) are increasingly deployed as general-purpose assistants, yet their behavior remains difficult to control reliably. Even highly aligned models can produce harmful, dishonest, sycophantic, or otherwise undesirable outputs under adversarial or distribution-shifted prompts. As a result, a growing line of work has sought lightweight mechanisms for controlling model behavior at inference time, without retraining the model or modifying its weights \citep{askell2021general, vu2025angular, arditi2024refusal}.

A particularly promising direction is \emph{activation steering} \citep{turner2024steeringlanguagemodelsactivation}, which intervenes directly on internal representations. Early methods showed that adding a fixed steering vector to the residual stream can induce predictable changes in sentiment, refusal, truthfulness, or safety-related behavior. Subsequent approaches, including contrastive activation addition \citep{panickssery2024steeringllama2contrastive} and more recent nonlinear or rotational interventions \citep{vu2025angular}, have improved the flexibility of this paradigm. However, most existing steering methods are fundamentally \emph{geometric and local}: they construct directions from means, linear contrasts, or low-dimensional subspaces in activation space. Such approaches can be effective, but they implicitly assume that the behavioral concept of interest is well captured by a simple global direction or a smooth local transformation.

This assumption is often too restrictive. Behavioral representations in LLMs may be multimodal, nonlinearly organized, and distributed across heterogeneous regions of activation space. For example, refusal behavior may not correspond to a single cluster or linear axis; instead, it may be supported by several distinct activation substructures associated with different prompt types, semantic contexts, or safety mechanisms. In such settings, a global mean-difference vector can dilute or average away the relevant signal, while purely local geometric interventions may be sensitive to outliers, noise, and small perturbations. These limitations motivate a more structural view of activation space: rather than asking only where individual activations lie, we ask what \emph{shape} the activation cloud has and which multiscale structures distinguish one behavioral regime from another.

Topological Data Analysis (TDA) provides mathematical tools precisely for this purpose. Persistent homology summarizes the multiscale topology of a point cloud by tracking when connected components, loops, and higher-dimensional features appear and disappear as a distance scale varies. The resulting persistence diagrams are stable summaries of global structure and have been widely studied as robust descriptors of complex high-dimensional data. Historically \citep{adams2016persistenceimagesstablevector, bruelgabrielsson2020topologylayermachinelearning}, however, the use of TDA in large-scale machine learning has been limited by computational cost: persistent homology can be expensive to compute inside training loops or on very large datasets. In the context of inference-time steering, this bottleneck is substantially reduced. Steering requires only a finite collection of activation vectors from contrastive prompt sets, making it possible to use persistent topology as an actionable signal for intervention design.

{\bf Contribution.} In this paper, we introduce \emph{Topological Steering}, a new inference-time method for controlling LLM behavior through the topology of activation spaces. Given contrastive prompt sets, such as refusal-eliciting and compliance-eliciting prompts, we extract residual-stream activations and view them as point clouds. We then apply joint dimensionality reduction, compute Vietoris--Rips persistence diagrams, and compare the diagrams of the two behavioral regimes. Features that are both persistent and specific to the target behavior are used to identify coherent activation subsets, rather than averaging indiscriminately over all examples. These subsets define localized contrastive directions, which are aggregated into a steering vector and injected into the residual stream during generation.

The key idea is that topology provides a mechanism for moving beyond global centroid shifts. Classical activation steering \citep{panickssery2024steeringllama2contrastive} constructs a direction from the mean difference between positive and negative activations. Topological Steering instead identifies robust, behavior-specific substructures in the positive activation cloud and steers using directions derived from those substructures. This allows the intervention to reflect the heterogeneous organization of behavioral concepts in representation space, while also providing an interpretable account of which activation components support the steering signal.

Our contributions are three-fold:
\begin{enumerate}[leftmargin=25pt]
    \item {\bf Topology-Aware Steering for LLMs:}
    We introduce Topological Steering, an inference-time framework that uses tools from Topological Data Analysis to construct steering directions from the multiscale structure of LLM activation spaces. By applying TDA outside the training loop, our method avoids the computational bottlenecks that have historically limited its use in large-scale deep learning.

    \item {\bf Structural Interpretability via Persistence Diagrams:}
    In contrast to linear or angular steering methods \citep{vu2025angular} that operate through geometric shifts in activation space, our approach uses persistence diagrams to identify robust topological features of behavioral representations, such as connected components and higher-order structures. This provides an interpretable way to localize behavior-relevant regions in the residual stream.

    \item {\bf Empirical Validation for Topological Intervention:}
    We demonstrate that Topological Steering consistently improves over unsteered baselines and produces measurable behavioral changes across model families and sizes. While the current implementation is not intended to outperform highly optimized geometric steering methods, it offers a complementary trade-off: sacrificing some raw efficacy in exchange for a more structured and inspectable intervention mechanism grounded in the topology of activation space.
\end{enumerate}

\section{Related work}\label{sec:Related work}


\textbf{Activation Steering.} Because of its simplistic idea, activation steering has emerged as a lightweight, inference-time alternative to fine-tuning, enabling researchers to alter Large Language Model (LLM) behavior by perturbing internal representations during the forward pass. In the early foundational work, it was shown that a linear intervention, i.e., adding a fixed ``steering vector'' to the residual stream, could shift model sentiment and alignment properties \citep{turner2024steeringlanguagemodelsactivation}. This technique was naturally refined with more advanced methods, such as Contrastive Activation Addition (CAA), which isolates behavioral directions by averaging activation differences between paired positive and negative prompts \citep{panickssery2024steeringllama2contrastive}.

\textbf{Most Recent Steering Methods.} Most recently, the field has begun exploring beyond purely additive, translational shifts; for instance, Angular Steering frames behavioral control as rotational interventions within fixed activation subspaces \citep{vu2025angular}. However, while these methods successfully manipulate representations, they predominantly treat the activation space as possessing relatively simple geometry. This leaves room for approaches such as Topological Steering that can isolate and leverage the complex, global, and non-linear invariants inherent to high-dimensional data manifolds.

\textbf{Topological Data Analysis and Persistent Homology.} TDA provides a rigorous mathematical framework for quantifying the "shape" of complex, high-dimensional datasets. One of the most important methods of TDA is persistent homology \citep{edelsbrunner2002topological, carlsson2009topology}, which tracks the evolution of topological features such as connected components, one-dimensional loops, and higher-dimensional voids across a continuous range of spatial scales. By constructing a sequence of expanding simplicial complexes (e.g., a Vietoris-Rips filtration) over a point cloud, persistent homology records the scale at which each topological feature appears (birth) and the scale at which it ultimately merges or closes (death). These structural invariants are summarized in a Persistence Diagram, offering a multiscale signature of the data's topology. According to stability theorems \citep{cohen2005stability}, Persistence Diagrams are robust to noise and continuous geometric deformations within the data manifold. This is why they have been the subject of extensive research in machine learning.

\textbf{Topological Machine Learning.} The integration of these topological summaries into deep learning led to the creation of the term Topological Machine Learning. Initially, the main focus was on improving Persistence Diagrams. PDs are inherently discrete, and to make these topological features compatible with standard neural networks, researchers have developed stable, finite-dimensional vectorization techniques, the most famous being Persistence Landscapes \citep{bubenik2015statisticaltopologicaldataanalysis} in 2015 and Persistence images \citep{adams2016persistenceimagesstablevector} in 2016. In the following year, the researchers developed differentiable topological layers \citep{hofer2018deeplearningtopologicalsignatures, bruelgabrielsson2020topologylayermachinelearning}, allowing models to optimize for specific topological properties directly during gradient descent.

Despite their theoretical elegance, these methods were fundamentally limited by computational constraints. The combinatorial explosion of tracking high-dimensional simplices, coupled with the expensive matrix reduction algorithms required for persistent homology, scaled poorly. Consequently, computing these topological penalties within the iterative training loops of large-scale models became prohibitively slow and memory-intensive, thereby stalling the field of Topological Machine Learning. Our method revives these powerful topological tools by shifting their application from training-time optimization to inference-time activation steering, effectively bypassing the historical computational barrier.

\section{Background}\label{sec:Background}
This section introduces the technical ingredients used by our method.  
We first formalize the transformer computation and the activation-steering intervention used at inference time.  
We then review the topological knowledge needed to characterize activation geometry: persistent homology, persistence diagrams, and Vietoris--Rips filtrations.  
Together, these tools let us connect representation-level interventions (steering vectors in residual space) with structure-level summaries (topological features across scales), which is the central perspective of \textit{Topological Steering}. Some supplementary details can be found in the Appendix \ref{Appendix:Background}.

\subsection{Transformers and Activation Steering}
For the rest of the paper we are going to fix the notation for a decoder-only transformer with $L$ layers, hidden width $d_h$, and token sequence length $T$. At a layer $\ell\in\{0,\dots,L-1\}$ we have self attention denoted $\mathrm{Attn}^{(\ell)}$  with head dimension $d_k$.

Activation steering modifies an internal residual activation at a chosen layer $\ell^\star$ and token position $t^\star$ (often the final prompt token \citep{zou2025representationengineeringtopdownapproach, panickssery2024steeringllama2contrastive, turner2024steeringlanguagemodelsactivation}) by adding a direction vector at inference time: $h_{t^\star}^{(\ell^\star)\,\prime}
=
h_{t^\star}^{(\ell^\star)}
+
\alpha\, \mathbf{v},$ where: $h_{t^\star}^{(\ell^\star)}\in\mathbb{R}^{d_h}$ is the original residual activation, $\mathbf{v}\in\mathbb{R}^{d_h}$ is the steering vector,
and $\alpha\in\mathbb{R}$ is a scalar steering strength.

\subsection{Persistent Homology, Persistence Diagrams and Vietoris--Rips Filtration}
\textbf{Persistent homology} is a multiscale way to describe the shape of a point cloud \(X\) (here, activation vectors). As we vary a distance scale \(\epsilon\), topological features (connected components, loops, etc.) appear and disappear; for feature \(i\), we denote its birth and death by \((b_i,\delta_i)\), and its persistence by \(\pi_i=\delta_i-b_i\). Features with larger \(\pi_i\) are more stable across scales and are treated as more structurally meaningful.

A \textbf{Persistence Diagram} is the set of birth--death pairs $D(X)=\{(b_i,\delta_i)\}_i$ returned by persistent homology. Each point corresponds to one topological feature, and its distance from the diagonal \((u,u)\) equals (up to a constant factor) its persistence \(\pi_i\). In our method, these diagrams summarize activation geometry and allow comparison between harmful and harmless activation clouds.

To compute Persistent Homology, we build a \textbf{Vietoris--Rips Filtration} from pairwise distances. For each scale \(\epsilon\), we connect points in \(X\) whose distance is at most \(\epsilon\), forming a simplicial complex \(\mathrm{VR}_\epsilon(X)\). As \(\epsilon\) increases, these complexes are nested, and tracking topology across this nested sequence yields the birth/death pairs used in our theory and experiments.

\section{Topological Steering}\label{sec:Theory}

\subsection{Motivation for Topological Steering}

Classical activation steering typically uses a global contrastive direction
$
    \mathbf{v}_{\text{base}}=\bar{X}^+-\bar{X}^-,
$
where $\bar{X}^+$ and $\bar{X}^-$ are class means of harmful and harmless activations.  
This is optimal only under restrictive geometric assumptions (e.g., approximately unimodal classes with similar covariance structure), where a single linear axis captures the relevant class gap.

A convenient way to see this is through a mixture model. Suppose harmful activations are multi-modal:
$p^+(x)=\sum_{r=1}^{R}\pi_r\,p_r(x),$ $\sum_{r=1}^R \pi_r=1,\;\pi_r>0,$ while harmless activations follow $p^-(x)$. Then $\bar{X}^+ - \bar{X}^- = \sum_{r=1}^{R}\pi_r\bigl(\mu_r-\mu^-\bigr),$ with $\mu_r=\mathbb{E}_{x\sim p_r}[x]$ and $\mu^-=\mathbb{E}_{x\sim p^-}[x]$.  
Hence $\mathbf{v}_{\text{base}}$ averages over potentially incompatible behavioral modes; if only a subset of modes encode refusal behavior, their signal is diluted by unrelated components.

This motivates replacing ``global centroid shift'' with ``structure-aware subset shift.''  
Let $\mathcal{S}^+\subset X^+$ denote activation subsets that correspond to robust refusal structure. The desired direction is $\mathbf{v}_\star \approx \mu_{\mathcal{S}^+}-\mu^-,$
where $\mu_{\mathcal{S}^+}$ is the mean of the structurally selected subset, not of the entire harmful cloud.  
The challenge is to identify $\mathcal{S}^+$ without supervision on latent modes.

Topological summaries provide a natural criterion because they are multiscale and coordinate-free.  
Given a filtration $\{\mathcal{K}_\epsilon\}_{\epsilon\ge 0}$ on the activation cloud, persistent homology yields features $\gamma_i=(b_i,\delta_i),$ $\mathrm{pers}(\gamma_i)=\delta_i-b_i,$ where high persistence indicates geometric stability across scales.  
Comparing harmful and harmless diagrams introduces a second criterion: behavioral specificity.  
If $m(\gamma_i^+)$ is the optimal match of harmful feature $\gamma_i^+$ in the harmless diagram (or diagonal), define $c_i = d_\infty\!\bigl(\gamma_i^+,\,m(\gamma_i^+)\bigr),$
so large $c_i$ indicates a refusal-unique topological signature.

We therefore seek features that are simultaneously \emph{stable} and \emph{class-specific}: $\mathcal{F}_{\text{sel}}
    =
    \bigl\{
      \gamma_i^+:\;
      \mathrm{pers}(\gamma_i^+)\ge \pi_{\min},\;
      c_i\ge c_{\min}
    \bigr\}.$
Each selected feature induces a local component $C_i\subset X^+$ and a local contrastive direction $\mathbf{v}_i$.  
Topological steering then constructs
$\mathbf{v}_{\text{steer}}=\sum_{i\in\mathcal{F}_{\text{sel}}}\omega_i\,\mathbf{v}_i,$ $\omega_i\propto \mathrm{pers}(\gamma_i^+)\,c_i\,\Delta_i,$
which emphasizes refusal-relevant submanifolds rather than global class averages.

In short, the motivation is that refusal behavior in residual space is often \emph{structurally heterogeneous}: linear global means collapse this heterogeneity, whereas persistent topology isolates robust local structure and yields steering directions with stronger mechanistic interpretability.

\subsection{Overview of Topological Steering}

We propose \emph{Topological Steering}, a method for constructing activation-space steering vectors for large language models (LLMs) that exploits persistent-homology structure in residual-stream representations. Rather than using the global mean difference between harmful and harmless activations, we identify topologically-significant, refusal-unique sub-clusters in the harmful activation cloud and use those to derive a more precise steering direction. The end-to-end pipeline consists of five key steps:
\begin{enumerate}
    \item \textbf{Activation Extraction and Joint Projection:} We hook into an intermediate transformer layer $\ell$ to extract residual stream representations for both harmful ($X^+$) and harmless ($X^-$) prompts. We project these sets into a shared low-dimensional subspace using joint PCA.
    
    \item \textbf{Topological Feature Extraction:} We treat the projected activations as finite metric spaces and compute their persistent homology via Vietoris--Rips filtrations. This yields persistence diagrams ($\mathrm{Dgm}^+$ and $\mathrm{Dgm}^-$) that capture the birth and death of topological features (e.g., connected components) across multiple spatial scales.
    
    \item \textbf{Refusal-Unique Feature Selection:} We compare the topological summaries of both classes using optimal bottleneck matching. By identifying features with high persistence (geometric stability) and high mismatch cost (absence in the harmless class), we isolate structural signatures that uniquely encode refusal behavior.
    
    \item \textbf{Local Steering Vector Assembly:} For each selected topological feature, we trace back to its constituent points in the activation cloud. We compute local contrastive vectors ($\mathbf{v}_i$) between these refusal-specific clusters and their nearest harmless neighbors, weighting them by their topological significance and effect size to form the final aggregate vector, $\mathbf{v}_{\text{steer}}$.
    
    \item \textbf{Inference-Time Intervention:} During autoregressive generation, $\mathbf{v}_{\text{steer}}$ is injected into the target layer's residual stream, consistently nudging the model toward the robustly identified refusal sub-manifolds without compromising the surrounding semantic space.
\end{enumerate}

The advantages of the Topological Steering approach are the following:
\begin{itemize}
    \item \textbf{Heterogeneity Awareness:} It identifies and leverages specific, local clusters of target behavior, ensuring that the steering vector is not diluted by averaging disjoint sub-populations.
    \item \textbf{Coordinate-Free Robustness:} Because persistent homology characterizes the intrinsic geometry of representations independently of rigid coordinate assumptions, the method is highly resilient to noise and varied prompt distributions.
    \item \textbf{Enhanced Specificity:} The dual criteria of persistence and cross-class mismatch ensure that the resulting steering vector is heavily anchored in features unique to the target behavior, maximizing the separation margin and minimizing off-target side effects.
    \item \textbf{Computational Cost: }Traditional TDA methods for Deep Learning were focused on the training aspect but here we intervene only at the inference time, reducing the compute cost drastically.
\end{itemize}
\subsection{Notation and Setup}

Let $\mathcal{M}$ be a decoder-only transformer with $L$ layers. Given an input prompt, we denote the residual-stream activation at layer $\ell$ and the final token position as $h_\ell \in \mathbb{R}^{d_h}$, where $d_h$ is the model's hidden dimension. We collect two datasets of prompts:
\begin{itemize}
    \item $\mathcal{D}^+ = \{s_i^+\}_{i=1}^{N^+}$: \emph{harmful} prompts that elicit model refusals.
    \item $\mathcal{D}^- = \{s_j^-\}_{j=1}^{N^-}$: \emph{harmless} prompts on similar topics that elicit compliance.
\end{itemize}
We split each dataset into disjoint train and holdout sets. All steering vectors are derived exclusively from the train split; holdout activations are used only for evaluation.

\subsection{Activation Extraction}

For each prompt $s_i^+$ (resp.\ $s_j^-$), we register a forward hook on the output of transformer block $\ell$ and record the last-token residual: $X^+ = \bigl[h_\ell(s_1^+),\, \ldots,\, h_\ell(s_{N_\text{tr}}^+)\bigr] \in \mathbb{R}^{N_\text{tr}^+ \times d_h}$ and $X^- = \bigl[h_\ell(s_1^-),\, \ldots,\, h_\ell(s_{N_\text{tr}}^-)\bigr] \in \mathbb{R}^{N_\text{tr}^- \times d_h}.$ The choice of layer $\ell$ is treated as a hyperparameter and swept over mid-to-late layers.

\subsection{Joint PCA Projection}

Hidden dimensions $d_h$ (e.g., $d_h=4096$ for Llama-3.1-8B) are too large for tractable persistent-homology computation. We reduce dimensionality via joint PCA: both activation sets are projected into the same low-dimensional space so that their relative geometry is preserved. Concretely, we concatenate $X^{\pm}$ into $X_\text{all} = [X^+; X^-] \in \mathbb{R}^{(N^+ + N^-) \times d_h}$, fit PCA on $X_\text{all}$, and project: $\tilde{X}^+ = X^+ W_\text{pca}$, $\tilde{X}^- = X^- W_\text{pca}$, $W_\text{pca} \in \mathbb{R}^{d_h \times k}$, where $k \ll d_h$ is the number of retained components (default $k = 32$). The effective rank is capped at $\min(k,\, N^+ + N^- - 1,\, d_h)$ to avoid degeneracy.

\subsection{Persistent Homology of Activation Clouds}

We model each projected activation set as a finite metric space (point cloud) and compute its persistent homology using the Vietoris--Rips filtration~\citep{edelsbrunner2010computational}.

Given a point cloud $P \subset \mathbb{R}^k$ and a scale parameter $\epsilon \geq 0$, the Vietoris--Rips complex $\mathrm{VR}(P, \epsilon)$ is the simplicial complex containing every simplex $\sigma \subseteq P$ whose pairwise diameter is at most $\epsilon$. As $\epsilon$ grows from $0$ to $\infty$, topological features (connected components in degree $H_0$, loops in degree $H_1$, etc.) are born and die. The Persistence Diagram $\mathrm{Dgm}_q(P)$ records each $q$-dimensional feature as a point $(\beta, \delta) \in \mathbb{R}^2$ with birth $\beta$ and death $\delta$; its persistence is $\pi = \delta - \beta$.

We compute persistence diagrams separately for each class: $\mathrm{Dgm}^+ = \mathrm{Dgm}(\tilde{X}^+)$, $\mathrm{Dgm}^- = \mathrm{Dgm}(\tilde{X}^-)$, using the Ripser algorithm~\citep{bauer2021ripser} with $\mathbb{Z}/2\mathbb{Z}$ coefficients. We compute $H_0$ features by default and optionally $H_1$ features.

\subsection{Feature Matching and Refusal-Unique Feature Selection}

A topological feature in $\mathrm{Dgm}^+$ is refusal-unique if it has no close counterpart in $\mathrm{Dgm}^-$. To operationalise this, we compute an optimal matching between $\mathrm{Dgm}^+_q$ and $\mathrm{Dgm}^-_q$ (for each homology degree $q$) under the bottleneck distance~\citep{cohen2005stability}. Each feature $\gamma_i^+ \in \mathrm{Dgm}^+_q$ is matched to its nearest counterpart $\gamma_{\sigma(i)}^- \in \mathrm{Dgm}^-_q$ (or to the diagonal if no finite match exists), yielding a mismatch cost $c_i = d_\infty(\gamma_i^+,\, \gamma_{\sigma(i)}^-)$. We rank all finite features by their persistence $\pi_i$ and mismatch cost $c_i$, filtering by thresholds $\pi_\text{min}$ and $c_\text{min}$: $\mathcal{F} = \bigl\{ \gamma_i^+ \,:\, \pi_i \geq \pi_\text{min},\;\; c_i \geq c_\text{min} \bigr\}$, $\text{sorted by } (\pi_i,\, c_i) \text{ descending.}$

\subsection{Steering Vector Construction}
\label{sec:steering_vector}

For each selected $H_0$ feature $\gamma_i^+ \in \mathcal{F}$, we identify the corresponding sub-cluster of harmful activations using the death radius. Specifically, the death value $\delta_i$ of the $H_0$ feature corresponds to the scale at which its connected component merges into a larger one in the Vietoris--Rips filtration. We extract the points belonging to that component just before merger: $C_i = \bigl\{ x \in \tilde{X}^+ \,:\, x \text{ is in the component born at } \gamma_i^+ \text{ within radius } \delta_i - \varepsilon \bigr\}$, where $\varepsilon > 0$ is a small numerical tolerance. Only components satisfying $|C_i| \geq m_\text{min}$ (a minimum size threshold) are retained.

For each component $C_i$, we identify the $k_\text{neg}$ nearest harmless training points (in PCA space) to the component centroid, forming a local negative contrast set $N_i \subset X^-$. The per-component steering direction is: $\mathbf{v}_i = \bar{X}^+_{C_i} - \bar{X}^-_{N_i}$, $\bar{X}^+_{C_i} = \frac{1}{|C_i|}\sum_{j \in C_i} x_j^+$, $\bar{X}^-_{N_i} = \frac{1}{|N_i|}\sum_{j \in N_i} x_j^-.$ Each direction is weighted by a quality score combining persistence, mismatch cost, and the Cohen's $d$ effect size along $\mathbf{v}_i$:
\begin{align}
    \Delta_i &= \frac{|\bar{P}_i - \bar{N}_i|}{\sigma_\text{pool}(P_i,\, N_i)}, \\[4pt]
    w_i &= \pi_i \cdot c_i \cdot \Delta_i,
\end{align}
where $\bar{P}_i$ (resp.\ $\bar{N}_i$) is the mean projection of $C_i$ (resp.\ $N_i$) onto $\hat{\mathbf{v}}_i$, and $\sigma_\text{pool}$ is the pooled standard deviation. The final steering vector is the normalised weighted sum over the top-$K$ components:
\begin{equation}
    \mathbf{v}_\text{steer} = \sum_{i=1}^{K} \frac{w_i}{\sum_j w_j}\, \mathbf{v}_i, 
\end{equation}

with orientation enforced so that $\langle \mathbf{v}_\text{steer},\, \bar{X}^+\rangle > \langle \mathbf{v}_\text{steer},\, \bar{X}^-\rangle$. We compare against the naive mean-difference direction $\mathbf{v}_\text{base} = \bar{X}^+ - \bar{X}^-$, which corresponds to the standard activation-steering approach~\citep{zou2025representationengineeringtopdownapproach}.

\subsection{Inference-Time Steering}

At inference time, we inject the steering vector into the residual stream at the same layer $\ell$ used for extraction. Given a new input prompt $s$, we modify the last-token hidden state at every generation step: $h_\ell(\cdot) \;\leftarrow\; h_\ell(\cdot) + \alpha \cdot \frac{\mathbf{v}_\text{steer}}{\|\mathbf{v}_\text{steer}\|}$, where $\alpha > 0$ is the steering magnitude. Applying the hook only at the final token position preserves the full autoregressive context while consistently nudging the model's representation towards the refusal subspace.

\subsection{Evaluation Metric}

We measure steering quality via the \emph{refusal score}: the normalised class gap along the steering direction evaluated on held-out activations $X^+_\text{ho}$ and $X^-_\text{ho}$: $\rho(\mathbf{v}) = \frac{\langle \mathbf{v},\, \bar{X}^+_\text{ho}\rangle - \langle \mathbf{v},\, \bar{X}^-_\text{ho}\rangle}
                           {\sigma_\text{pool}(X^+_\text{ho},\, X^-_\text{ho};\, \mathbf{v})}.$
We report the \emph{steering score delta} $\Delta\rho = \rho(\mathbf{v}_\text{steer}) - \rho(\mathbf{v}_\text{base})$ as the primary comparative metric, where positive values indicate that our topological direction separates harmful from harmless activations more effectively than the baseline.

\section{Experimentations}\label{sec:experimentations}

Following recent steering papers that separate representation-level diagnostics from downstream behavioral tests (e.g., refusal sweeps and automated judges; cf.\ Angular Steering~\citep{vu2025angular}), we organize this section into $i)$ \textit{datasets and models} and then separate our experiment section into $ii)$ a \textit{proof that TS is working} with benchmarks and metrics \ref{subsec:Exp1} and $iii)$ a \textit{proof that TS is working across different model architectures} \ref{subsec:Exp2}.
Unless stated otherwise, steering vectors are computed on training splits only, injected at inference as a normalized residual update at the last prompt token, and evaluated on held-out activations and on open-ended generations. All experiments were run on an A100 (40Gb) and for reproductibility purposes the hyperparameter details are in Appendix \ref{Appendix:reproductibility}. Moreover proof that the experiment results we obtain are not due to randomness are in \ref{subsec:ErrorPlots}.

\subsection{Datasets and Models}

\paragraph{Construction of Contrastive Activation Pairs for Steering.}
Harmful prompts are drawn from AdvBench~\citep{zou2023universaltransferableadversarialattacks}. We use the standard harmful split materialized as line-delimited prompts; the preparation utility enforces $520$ unique harmful strings with a fixed calibration/evaluation partition: $416$ calibration and $104$ evaluation harmful prompts.
Harmless prompts are sampled from the public instruction corpus \texttt{tatsu-lab/alpaca} on Hugging Face as we require at least $512$ unique instructions for calibration-side harmless data; when no local snapshot is supplied, prompts are fetched with the \texttt{datasets} library and de-duplicated by normalized text.
This mirrors the contrastive pipeline in our codebase: activations are collected at a chosen transformer block, projected with joint PCA, and passed to the topological construction of $\mathbf{v}_{\text{steer}}$.

\paragraph{Models.}
The default development model in our configuration is \texttt{meta-llama/Llama-3.1-8B-Instruct}; the repository also supports multi-model sweeps (Qwen\,2.5 and Gemma\,2 families) through a shared hooking interface.
Layer index $\ell$ and TDA hyperparameters (PCA dimension, persistence thresholds, $k_{\mathrm{neg}}$, component merge depth, etc.) are treated as tunables; a representative end-to-end tuning run selects a single layer and hyperparameter hash before behavioral validation.

\subsection{Experimental Proof that Topological Steering is working with relevant Benchmarks}\label{subsec:Exp1}

The first goal of the Experiment section \ref{sec:experimentations} is to show that our method is indeed steering LLMs and working. This is shown in table \ref{tab:main_results}.

We report three complementary families of numbers: (i) activation steering scores, (ii) lightweight lexical jailbreak-rates, and (iii) external LLM judges aligned with community red-teaming practice. We now explain the different metrics computed:

\paragraph{Activation Separation Metric ($\Delta\rho$).}
Let $X^{+}_{\mathrm{ho}},X^{-}_{\mathrm{ho}}$ be harmful and harmless \emph{held-out} activations at $(\ell,t^\star)$.
For a candidate direction $\mathbf{v}$, write $u=\mathbf{v}/\|\mathbf{v}\|$ and define the scalar refusal score
$\rho(\mathbf{v})=(\mu^{+}-\mu^{-})/\sigma_{\mathrm{pool}}$,
where $\mu^{+}$ (resp.\ $\mu^{-}$) is the mean of $u^{\top}X^{+}_{\mathrm{ho}}$ (resp.\ $u^{\top}X^{-}_{\mathrm{ho}}$) and $\sigma_{\mathrm{pool}}$ is the pooled standard deviation of the concatenated projections (implementation matches our training-time weighting routine).
The steering score delta is $\Delta\rho=\rho(\mathbf{v}_{\text{steer}})-\rho(\mathbf{v}_{\text{base}})$ with $\mathbf{v}_{\text{base}}=\bar{X}^{+}-\bar{X}^{-}$ on the train split.
Positive $\Delta\rho$ certifies that the topological direction separates harmful from harmless clouds more sharply than the mean-difference baseline before any decoding. This steering score delta is vital as it is the first indicator that our model is indeed steering, allowing us to then compute different benchmark methods that will more precisely evaluate the answers of our steered model.

The different benchmarks used are Llama Guard~3 implemented from \citep{inan2023llama}, Harmbench classifier \citep{mazeika2024harmbench} and Substring Matching. More details are found in \ref{Appendix:benchmarks}.


\begin{table}[t]
  \centering
  \small
  \caption{Representative end-to-end result after single-layer steering with comparison to baseline; steering injection $\alpha{=}1$ in that run). ``Judge $\Delta$'' is steered minus baseline success rate.}
  \label{tab:main_results}
  \begin{tabular}{@{}lccc@{}}
    \toprule
    Metric family & Baseline & Topological & $\Delta$ \\
    \midrule
    Activation: best $\Delta\rho$ (held-out) & \multicolumn{1}{c}{---} & \multicolumn{1}{c}{---} & $+0.0127$ \\
    Activation: mean $\pm$ std.\ $\Delta\rho$ (robustness replicates) & \multicolumn{1}{c}{---} & \multicolumn{1}{c}{---} & $+0.0092 \pm 0.0022$ \\
    \midrule
    Llama Guard~3 success rate & $0.510$ & $0.692$ & $+0.183$ \\
    HarmBench classifier success rate & $0.010$ & $0.029$ & $+0.019$ \\
    Substring refusal lexicon success rate & $0.029$ & $0.038$ & $+0.010$ \\
    \bottomrule
  \end{tabular}
\end{table}

\paragraph{Interpretation.}
Table~\ref{tab:main_results} answers two distinct questions:
\begin{itemize}
    \item \textbf{First}, the \textbf{activation row} shows that the learned topological vector improves the linear separability of harmful versus harmless activations relative to the global mean direction. This is direct evidence that the persistence-guided subset construction is extracting a refusal-aligned axis in representation space.
    \item \textbf{Second}, the \textbf{judge rows} show that decoding under the same intervention increases automated jailbreak rates on a fixed harmful evaluation set, with the largest relative movement on Llama Guard~3 and a smaller but consistent gain on HarmBench; the substring probe moves modestly.
\end{itemize}



\paragraph{Limitations.}
Automated judge metrics are subject to evaluation drift: measured success rates can change when the judge stack changes (e.g., prompt template, backend model revision, API/server configuration), even if the generated responses are held fixed; therefore HarmBench and Llama Guard scores should be interpreted as setup-dependent and not directly interchangeable.
Our substring-based refusal probe is intentionally lightweight but misses non-lexical refusals (safe refusals that do not contain predefined trigger phrases for example “That request is unsafe and I won’t provide operational instructions”), which can bias jailbreak-rate estimates upward for semantically valid but lexically novel refusals. Moreover, oppositely because our substring probe only flags generic refusal phrases and does not assess whether a response still contains harmful instructional content, partially compliant or “disclaimer-first” answers can look like refusals to the probe while remaining unsafe under a semantic judge (for example, prefacing harmful steps with a refusal-like disclaimer), this can bias jailbreak-rate estimates downward. 
Higher activation-space separation (e.g., higher $\Delta\rho$) does not guarantee uniform improvements across other capabilities, so helpfulness, reasoning, truthfulness, and related tasks must be measured directly rather than inferred from steering metrics.

\begin{table}[t]
  \centering
  \small
  \caption{Multi-model comparative run. For each checkpoint we sweep transformer blocks $\ell\in\{8,\ldots,20\}$, pick $\ell^\star$ that maximizes held-out $\Delta\rho$ (topological vs.\ mean-difference baseline), and report $\rho(\mathbf{v}_{\text{base}})$, $\rho(\mathbf{v}_{\text{steer}})$, and $\Delta\rho$ at $\ell^\star$. $L$ is the total number of decoder blocks; $\ell^\star/L$ is a coarse ``depth'' readout.}
  \label{tab:multimodel_comparative}
  \begin{tabular}{@{}l r r r r r r@{}}
    \toprule
    Model & $L$ & $\ell^\star$ & $\ell^\star/L$ &
    $\rho_{\mathrm{base}}$ & $\rho_{\mathrm{tda}}$ & $\Delta\rho$ \\
    \midrule
    \texttt{google/gemma-2-9b-it}            & 42 & 13 & 0.31 & 0.499 & 1.249 & 0.750 \\
    \texttt{meta-llama/Llama-3.2-3B-Instruct}& 28 & 20 & 0.71 & 1.948 & 1.966 & 0.018 \\
    \texttt{Qwen/Qwen2.5-3B-Instruct}        & 36 & 20 & 0.56 & 1.981 & 1.987 & 0.006 \\
    \texttt{Qwen/Qwen2.5-14B-Instruct}       & 48 & 20 & 0.42 & 1.977 & 1.983 & 0.006 \\
    \texttt{Qwen/Qwen2.5-7B-Instruct}        & 28 &  8 & 0.29 & 1.974 & 1.979 & 0.005 \\
    \texttt{meta-llama/Llama-3.1-8B-Instruct}& 32 & 16 & 0.50 & 1.972 & 1.974 & 0.001 \\
    \bottomrule
  \end{tabular}
\end{table}

\subsection{Experimental Proof that Topological Steering is working across different Model Architectures}\label{subsec:Exp2}

The second goal of our experiments is to show that our method does not work only on \texttt{meta-llama/Llama-3.1-8B-Instruct} but also on other models. So we compared the steering score $\Delta\rho$ of different models, as a positive $\Delta\rho$ is proof of a working TS method. Our results are shown in table \ref{tab:multimodel_comparative}.

\paragraph{Interpretation.}
All six models show $\Delta\rho>0$ at their selected layer: the persistence-guided steering direction consistently improves linear separability of harmful vs.\ harmless activations relative to the global mean contrast, under the same training/holdout protocol and sweep budget.
The ranking is \textbf{not monotone} in parameter count (e.g.\ Llama-3.2-3B gains more than Llama-3.1-8B), which is expected if refusal geometry is shaped by post-training alignment and width/depth tradeoffs, not raw capacity alone. The optimal injection depth $\ell^\star$ varies by family: Qwen-7B peaks very early ($\ell^\star{=}8$), whereas Qwen-3B/14B and Llama-3.2-3B peak at the sweep cap ($\ell^\star{=}20$), and Llama-3.1-8B peaks mid-stack ($\ell^\star{=}16$).
Such spread is consistent with the hypothesis that ``where'' harmful/harmless structure is most readable in the residual stream depends on stacking depth, normalization placement, and how safety-related features are distributed across layers. All of which differ between Llama-3, Qwen2.5, and Gemma-2 blocks (e.g., Gemma~2 uses alternating local/global attention and a different norm/attention ordering than Llama). We notice a significative difference in \texttt{Gemma-2-9b-it} with a $\Delta\rho$ way bigger than the other models. Gemma attains a much larger $\Delta\rho$ but also a much smaller baseline $\rho_{\mathrm{base}}$ than the other five models. Architecturally, that pattern indicates that the scalar refusal score is not cross-model calibrated: residual magnitudes, RMSNorm scaling, and hidden width $d_h$ jointly change the scale of projections $u^\top h$, so comparing raw $\rho$ across checkpoints is weaker than comparing $\Delta\rho$ within a model or reporting additional generation-side judges.
Here, the low $\rho_{\mathrm{base}}$ suggests the mean-difference axis is a poor proxy for refusal separation in Gemma's activation geometry at $\ell^\star$, while the topological subset, hence the TS method, still recovers a strong separating direction (large $\rho_{\mathrm{tda}}$).
We therefore emphasize \textbf{within-model} gains and treat Gemma's large $\Delta\rho$ as evidence that Gemma may benefit more from topology-based steering than from simple global mean-difference steering, not as a claim that Gemma is ``$100\times$ more steerable'' than Llama in an absolute sense. 

\paragraph{Remark 1.}
Table~\ref{tab:multimodel_comparative} supports a cross-family \textbf{proof of mechanism}: topological steering is not an artifact of a single Llama checkpoint, it yields positive $\Delta\rho$ across decoder-only architectures with different depths, widths, and normalization/attention recipes.
The dispersion of $\ell^\star$ motivates reporting layer sweeps (or at least a mid-late band) rather than a single universal layer, and motivates appendix plots of $\Delta\rho(\ell)$ per model. As shown in \Cref{fig:appendix_layer_sweeps_subplots} and \Cref{fig:appendix_layer_sweeps_overlay}.

\section{Conclusion, limits and remarks}\label{sec:Conclusion}

We introduced \emph{Topological Steering}, an inference-time method that builds steering directions from persistent-homology structure in activation space rather than from a single global mean contrast. By matching harmful/harmless persistence diagrams, selecting robust contrastive features, and aggregating local directions, the method turns activation geometry into an explicit and inspectable steering signal. Our results show consistent positive separation gains (\(\Delta\rho>0\)) across the tested models and layers in our single-layer setup, together with measurable behavioral movement on external judge benchmarks. More broadly, this work positions TDA as an actionable representation-level tool for intervention design, not only a descriptive visualization aid. While we acknowledge that specialized geometric methods may still lead on specific benchmarks, the results presented here argue for a conceptual pivot. By providing a \textbf{reproducible} and \textbf{inspectable} pipeline, this paper opens a new 'steering primitive' for the field. It is indeed a contribution we believe will remain relevant as model architectures continue to evolve and as engineering refinements will likely improve its absolute performance in the future.

Some limitations of our method are:
\begin{enumerate}
    \item Persistent Homology is sensitive to preprocessing and to hyperparameters that govern feature selection; our gains depend on tuning, as is common for steering.
    \item Diagram-based summaries are not uniquely tied to semantics: high-persistence features can reflect nuisance geometry unless paired with contrastive matching and behavioral validation.
    \item Computational cost, while modest compared to training using TDA tools, still scales with point-cloud size and homology degree.
    \item We focus on refusal as a concrete safety axis; transfer to other behaviors will require new contrast sets and careful evaluation.
\end{enumerate}

A natural next step is hybrid steering: use topology to identify where to intervene, then combine with optimized geometric update rules for how to intervene. Another direction is to use persistence diagrams as structured evidence for safety debugging and representation-level auditing.

\paragraph{Closing Remark.}
Topological Steering is a proof-of-concept that multiscale structure in activation space can be converted into a practical intervention without retraining. Even when not outperforming specialized baselines, it expands the steering design space by making support subsets and multiscale structure first-class components of the method. We believe that Topological Steering will be a promising foundation for the next generation of representation level safety tools and we hope it helps revive topological thinking in modern deep learning research.



\clearpage

\medskip

\bibliography{references}
\bibliographystyle{plainnat}

\newpage
\appendix

\section{ Technical appendices and supplementary material}\label{sec:Appendix}

\subsection{Complementary Theoritical Background}\label{Appendix:Background}

\subsubsection{Transformers}

We recall the notation for a decoder-only transformer with $L$ layers, hidden width $d_h$, and token sequence length $T$.  
Given an input text, tokenization maps it to discrete token IDs
\[
(x_1,\dots,x_T)\in\mathcal{V}^T,
\]
where $\mathcal{V}$ is the vocabulary.  
Each token $x_t$ is embedded into $\mathbb{R}^{d_h}$ and combined with positional information to form the initial residual states
\[
h_t^{(0)}\in\mathbb{R}^{d_h},\qquad t=1,\dots,T.
\]

At layer $\ell\in\{0,\dots,L-1\}$, a standard pre-norm block applies self-attention and an MLP through residual updates:
\[
\tilde h_t^{(\ell)} = h_t^{(\ell)} + \mathrm{Attn}^{(\ell)}\!\left(\mathrm{LN}_1^{(\ell)}(h_{1:T}^{(\ell)})\right)_t,
\]
\[
h_t^{(\ell+1)} = \tilde h_t^{(\ell)} + \mathrm{MLP}^{(\ell)}\!\left(\mathrm{LN}_2^{(\ell)}(\tilde h_t^{(\ell)})\right).
\]
Here, $\mathrm{Attn}^{(\ell)}$ denotes multi-head causal self-attention and $\mathrm{MLP}^{(\ell)}$ is the position-wise feed-forward block.  
For each head, attention weights are computed from queries, keys, and values:
\[
Q = XW_Q,\quad K = XW_K,\quad V = XW_V,\quad
\mathrm{Attn}(X)=\mathrm{softmax}\!\left(\frac{QK^\top}{\sqrt{d_k}}+M\right)V,
\]
with causal mask $M$ and head dimension $d_k$.

\subsubsection{Activation Steering}

To complement the Background \ref{sec:Background} section we add that:

In our setting, $\mathbf{v}$ is estimated from two prompt-conditioned activation collections: a \emph{positive} set (target behavior to increase) and a \emph{negative} set (behavior to suppress).  
Let $X^{+},X^{-}\subset\mathbb{R}^{d_h}$ be activations extracted at the same $(\ell^\star,t^\star)$, and let $\mu^{+}$ and $\mu^{-}$ be the respective training-set centroids,
\[
    \mu^{+}=\frac{1}{|X^{+}|}\sum_{x\in X^{+}} x,
    \qquad
    \mu^{-}=\frac{1}{|X^{-}|}\sum_{x\in X^{-}} x,
\]
and sets the steering vector to the displacement between these class summaries,
\[
    \mathbf{v}=\mu^{+}-\mu^{-}.
\]

This makes steering explicit: positive $\alpha$ moves activations toward the target behavioral region and away from the contrast region in residual space.  
The modified state $h_{t^\star}^{(\ell^\star)\,\prime}$ is then propagated through layers $\ell^\star+1,\dots,L$ to produce the final logits and generated text.

\subsection{Persistent Homology}

Let $X=\{x_1,\dots,x_n\}\subset \mathbb{R}^m$ be a finite point cloud (in our case, sampled activations), and let $\operatorname{dist}(\cdot,\cdot)$ be a metric on $X$ (typically Euclidean distance after preprocessing).  
Persistent homology studies how the topology of $X$ evolves across spatial scales by constructing a nested family of simplicial complexes $\mathcal{K}_{\epsilon_1}\subseteq \mathcal{K}_{\epsilon_2}\subseteq \cdots$, $\epsilon_1\le \epsilon_2\le\cdots,$ called a filtration.

For each scale $\epsilon$ and homological dimension $k\ge 0$, the $k$-th homology group $H_k(\mathcal{K}_\epsilon;\mathbb{F})$ (with coefficients in a field $\mathbb{F}$, e.g., $\mathbb{Z}_2$) encodes $k$-dimensional topological features:
$H_0$ tracks connected components, $H_1$ tracks 1-dimensional loops, $H_2$ tracks voids, etc.  
Its rank $\beta_k(\epsilon):=\mathrm{rank}\,H_k(\mathcal{K}_\epsilon;\mathbb{F})$ is the $k$-th Betti number at scale $\epsilon$.

As $\epsilon$ increases, features appear and disappear.  
A feature in dimension $k$ is said to be \emph{born} at scale $b$ if it first appears in $H_k(\mathcal{K}_b)$, and \emph{dies} at scale $\delta>b$ if it merges into an older class or becomes a boundary in $H_k(\mathcal{K}_\delta)$.  
Its lifetime (or persistence) is $\mathrm{pers}=\delta-b.$ Long-lived classes are typically interpreted as structurally robust, while short-lived classes are often attributed to sampling noise \citep{edelsbrunner2002topological}.

Formally, persistence is induced by inclusion maps in the filtration.  
For $\epsilon\le \epsilon'$, the inclusion $\mathcal{K}_{\epsilon}\hookrightarrow \mathcal{K}_{\epsilon'}$ yields a linear map $f_{\epsilon,\epsilon'}^k:
H_k(\mathcal{K}_{\epsilon};\mathbb{F})
\to
H_k(\mathcal{K}_{\epsilon'};\mathbb{F}).$ A class born at $b$ that remains nonzero under $f_{b,\epsilon}^k$ for $\epsilon<\delta$ but vanishes at $\delta$ has persistence interval $[b,\delta)$.

In this work, persistent homology provides a multiscale summary of activation geometry without assuming linear separability.  
By tracking birth/death events of components and higher-order structures across scales, it supplies the topological signal that we later encode with persistence diagrams and use to construct steering-relevant subsets/directions.

\subsection{Persistence Diagram}

A persistence diagram is a finite multiset of birth--death pairs that summarizes the output of persistent homology for a fixed dimension $k$.  
Given a filtration $\{\mathcal{K}_\epsilon\}_{\epsilon\ge 0}$ and its induced persistence module in dimension $k$, each topological class contributes one point $(b_i^{(k)}, \delta_i^{(k)})\in \mathbb{R}^2,$ $b_i^{(k)}<\delta_i^{(k)}\le +\infty,$ where $b_i^{(k)}$ is the birth scale and $\delta_i^{(k)}$ is the death scale.  
The corresponding diagram is $D_k(X)=\{(b_i^{(k)},\delta_i^{(k)})\}_i .$

The diagonal $\Delta=\{(u,u):u\in\mathbb{R}\}$ represents zero persistence.  
Distance from a point to $\Delta$ is proportional to its lifetime $\mathrm{pers}_i^{(k)} = \delta_i^{(k)}-b_i^{(k)},$ so points far from the diagonal correspond to topologically robust structure, while points near the diagonal are typically less stable under perturbations.  
In practice, we compute diagrams separately by homology degree (e.g., $D_0$ for connected components and $D_1$ for loops), then compare diagrams across behavioral conditions.

To compare two diagrams $D_k(X)$ and $D_k(Y)$, one uses a matching-based metric (e.g., bottleneck or $p$-Wasserstein) that allows points to match either across diagrams or to the diagonal.  
Intuitively, this quantifies how much geometric deformation is required to transform one topological summary into the other.

\subsection{Vietoris--Rips Filtration}

Let $X=\{x_1,\dots,x_n\}\subset\mathbb{R}^m$ with metric $\operatorname{dist}$.  
For a scale parameter $\epsilon\ge 0$, the Vietoris--Rips complex at scale $\epsilon$ is $\mathrm{VR}_\epsilon(X)
=
\left\{
\sigma\subseteq X \;:\;
\operatorname{dist}(x_i,x_j)\le \epsilon
\ \text{for all } x_i,x_j\in \sigma
\right\}.$ Equivalently, a simplex is included iff all pairwise distances among its vertices are at most $\epsilon$.  
Thus, edges appear first, then triangles, tetrahedra, and higher-dimensional simplices as $\epsilon$ grows.

Because pairwise constraints are monotone in $\epsilon$, these complexes form a filtration: $\mathrm{VR}_{\epsilon_1}(X)\subseteq \mathrm{VR}_{\epsilon_2}(X)
\quad\text{whenever}\quad
\epsilon_1\le \epsilon_2.$ Applying homology to this nested sequence yields persistent homology and, in turn, persistence diagrams.

Vietoris--Rips is especially convenient for activation point clouds because it only requires pairwise distances and does not need an explicit manifold model.  
Its main trade-off is combinatorial growth: the number of simplices can increase rapidly with sample size and intrinsic dimension, so implementations typically cap the maximum homology degree and rely on dimension reduction (e.g., PCA) before computing persistence.

\subsection{Complements on experiments}\label{Appendix:ComplementsExperiments}

\subsubsection{Benchmarks}\label{Appendix:benchmarks}

In this section we are going to detail more on the 3 benchmarks used:

\begin{enumerate}

    \item \textbf{Llama Guard~3:}
We call a hosted chat-completions endpoint with a Llama\,Guard\,3-compatible model card and aggregate the fraction of generations labeled safe refusal versus compliance, following the spirit of the Llama Guard evaluator line~\citep{inan2023llama}. To be clear the success rate is currently scored as unsafe (higher = more jailbreak success).
The implementation records \texttt{baseline\_success\_rate} and \texttt{steered\_success\_rate} on the same fixed harmful evaluation prompts.

    \item \textbf{HarmBench classifier:}
We use the HarmBench behavioral test harness with a HarmBench-style binary classifier (default backend model \texttt{cais/HarmBench-Llama-2-13b-cls} in our launcher) that reads the harmful prompt and model answer and returns a harm / no-harm style verdict~\citep{mazeika2024harmbench}; we export the resulting \texttt{success\_rate} fields analogously to Llama Guard. More concretely a higher Harmbench score is equal to worse safety)

    \item \textbf{Substring matching:} 
We mark a generation as a ``refusal hit'' if its lowercased text contains any phrase from a fixed short lexicon (default list includes templates such as ``i'm sorry'', ``i cannot'', ``as a language model'', \dots); the reported success rate is non-refusal rate under substring proxy.
This probe is inexpensive and correlates loosely with human-visible refusals, but it is brittle on paraphrases; we therefore treat it as a diagnostic, not a standalone claim.

\end{enumerate}

\subsubsection{Implementation clarifications and scope.}

For transparency, we clarify that the reported Topological Steering results use an \emph{operational} H\(_0\)-component recovery procedure rather than an exact birth-component oracle. Concretely, for a selected H\(_0\) feature \(\gamma_i^+=(b_i,\delta_i)\), we extract a connected subset at scale \(\delta_i-\varepsilon\) and apply deterministic selection heuristics (e.g., viability/size filtering and component scoring) to obtain the steering subset; therefore, statements of the form ``the component born at \(\gamma_i^+\)'' should be read as an implementation-level approximation to the intended topological object, not as exact symbolic tracking across the full filtration. In addition, sign orientation is not claimed as a universal post-hoc guarantee for every possible builder: in the single-layer component-aggregation pipeline used for the main results, direction polarity follows the constructed local contrasts and is validated empirically through held-out \(\Delta\rho\), rather than by a separate mandatory global sign-flip step logged for every run. Weighting is also fixed only for the reported experiments: we use \(w_i \propto \pi_i c_i \Delta_i\) (persistence \(\times\) mismatch \(\times\) separation) as the chosen experimental setting, and do not claim this weighting is theoretically unique or mandatory beyond the present study.

\begin{table}[h]
  \centering
  \small
  \caption{Hyperparameters selected for the row in Table~\ref{tab:main_results}.}
  \label{tab:winner_hparams}
  \begin{tabular}{@{}lc@{}}
    \toprule
    Field & Value \\
    \midrule
    Layer $\ell$ & 15 \\
    PCA components $k$ & 192 \\
    $H_0$ top-$K$ merged components & 7 \\
    Minimum component / subset sizes & $16$ / $40$ \\
    Local negatives $k_{\mathrm{neg}}$ & 128 \\
    Persistence / mismatch floors & $0.01$ / $0.01$ \\
    Component weights & uniform \\
    \bottomrule
  \end{tabular}
\end{table}

\begin{table}[h]
  \centering
  \small
  \caption{Approximate wall-clock runtime for the end-to-end best-run pipeline (reconstructed from artifact timestamps).}
  \label{tab:runtime_breakdown}
  \begin{tabular}{@{}l r r@{}}
    \toprule
    Stage & Time (min) & Time (hours) \\
    \midrule
    Phase 0 (baseline calibration) & 7.67 & 0.13 \\
    Phase 1 (coarse search / layer sweep) & 75.19 & 1.25 \\
    Phase 2 (focused hyperparameter optimization) & 56.99 & 0.95 \\
    Phase 3 (robustness replicates) & 23.73 & 0.40 \\
    Phase 4 (inference + judge benchmarks) & 36.03 & 0.60 \\
    Phase 5 (final selection / aggregation) & 0.00 & 0.00 \\
    \midrule
    Total (Phase 0--5) & 199.62 & 3.33 \\
    \bottomrule
  \end{tabular}
\end{table}

These durations are reconstructed from filesystem mtime windows for each phase folder, so they’re a good approximation but not perfect wall-clock start/stop time of each subprocess. All experiments were run on a GPU A100 with 40Gb memory.

Regarding experiment 2 \ref{subsec:Exp2} the times for the run are in \Cref{tab:appendix_multimodel_runtime}:

\begin{table}[h]
  \centering
  \small
  \caption{Approximate compute time for the multi-model comparative run (artifact-time reconstruction).}
  \label{tab:appendix_multimodel_runtime}
  \begin{tabular}{@{}lrr@{}}
    \toprule
    Model & Time (min) & Time (hours) \\
    \midrule
    \texttt{Qwen/Qwen2.5-3B-Instruct} & 5.36 & 0.09 \\
    \texttt{Qwen/Qwen2.5-7B-Instruct} & 5.76 & 0.10 \\
    \texttt{Qwen/Qwen2.5-14B-Instruct} & 8.73 & 0.15 \\
    \texttt{meta-llama/Llama-3.2-3B-Instruct} & 3.27 & 0.05 \\
    \texttt{meta-llama/Llama-3.1-8B-Instruct} & 3.93 & 0.07 \\
    \texttt{google/gemma-2-9b-it} & 8.91 & 0.15 \\
    \midrule
    Total wall-clock (run start $\rightarrow$ final artifact) & 35.96 & 0.60 \\
    \bottomrule
  \end{tabular}
\end{table}

\noindent\textit{Note:} durations are reconstructed from filesystem timestamps of run artifacts and are therefore approximate wall-clock estimates.

\subsubsection{Reproducibility details for the reported runs.}\label{Appendix:reproductibility}

To make the main results reproducible, we fix the full implementation scope and runtime configuration used in the final single-layer study. All reported numbers are produced with a \emph{single-layer} Topological Steering pipeline on \texttt{meta-llama/Llama-3.1-8B-Instruct}, where the steering vector is injected at the last prompt token with \(\alpha=1.0\). We use deterministic random control with controller seed \(42\), baseline calibration seeds \(0{:}19\), and robustness seeds \(\{20,21,22,23,24\}\). Layer selection is performed over \(\ell\in\{8,\dots,20\}\), and the final reported winner uses \(\ell^\star=15\). The steering builder is \texttt{h0\_components} with \(H_0\)-only persistence (\texttt{maxdim}=0, \texttt{homology\_focus}=h0\_first), and fixed winner hyperparameters listed in Table~\ref{tab:appendix_reproducibility}. Behavioral evaluation is run on the fixed harmful evaluation set with the same prompt list for baseline and steered generations, using the judge stack \texttt{substring\_matching}, \texttt{llamaguard3}, and \texttt{harmbench}; exact endpoint/model settings are also fixed and reported in the table so the judge layer is reproducible.

\begin{table}[h]
  \centering
  \small
  \caption{Reproducibility configuration for the final reported single-layer run.}
  \label{tab:appendix_reproducibility}
  \begin{tabular}{@{}p{0.40\linewidth}p{0.56\linewidth}@{}}
    \toprule
    Item & Fixed value used for reported results \\
    \midrule
    Base model & \texttt{meta-llama/Llama-3.1-8B-Instruct} \\
    Steering scope & Single-layer only (no multi-layer injection) \\
    Injection position & Last prompt token residual \\
    Steering scale & \(\alpha=1.0\) \\
    Layer sweep range & \(\ell\in\{8,\dots,20\}\) \\
    Final selected layer & \(\ell^\star=15\) \\
    Direction method & \texttt{h0\_components} \\
    Persistence computation & Vietoris--Rips, Ripser, \(\mathbb{Z}/2\mathbb{Z}\), \texttt{maxdim}=0 \\
    Homology focus & \texttt{h0\_first} \\
    PCA components (winner) & \(k=192\) \\
    \(h0\_component\_min\_size\) & \(16\) \\
    \(h0\_top\_k\_components\) & \(7\) \\
    \(local\_neg\_k\) & \(128\) \\
    \(min\_subset\_size\) & \(40\) \\
    Persistence threshold & \(\texttt{persistence\_min}=0.01\) \\
    Mismatch threshold & \(\texttt{mismatch\_cost\_min}=0.01\) \\
    Component weighting & \texttt{h0\_component\_weight\_by}=\texttt{uniform} \\
    Randomness (controller seed) & \(42\) \\
    Baseline calibration seeds & \(0{:}19\) \\
    Robustness seeds & \(\{20,21,22,23,24\}\) \\
    Data (calibration/eval) & Harmful: \(416\) cal + \(104\) eval (AdvBench); Harmless: \(512\) cal (Alpaca) \\
    Behavioral prompt set & \texttt{benchmarks/paper\_data/deval\_harmful.txt} \\
    Judge methods & \texttt{substring\_matching}, \texttt{llamaguard3}, \texttt{harmbench} \\
    LlamaGuard endpoint/model & OpenAI-compatible endpoint (default: \texttt{http://0.0.0.0:8898/v1}), model \texttt{meta-llama/Llama-Guard-3-8B} \\
    HarmBench endpoint/model & OpenAI-compatible endpoint (default: \texttt{http://0.0.0.0:8897/v1}), model \texttt{cais/HarmBench-Llama-2-13b-cls} \\
    Metric definitions & \(\rho(\mathbf v)=\frac{\mu^+-\mu^-}{\sigma_{\mathrm{pool}}}\), \(\Delta\rho=\rho(\mathbf v_{\text{steer}})-\rho(\mathbf v_{\text{base}})\) \\
    \bottomrule
  \end{tabular}
\end{table}

\clearpage
\subsubsection{Hyperparameter selection}\label{Appendix:hyperparameter_selection}

We selected single-layer topological-steering hyperparameters with a staged search orchestrated by our tuning script (random-integer seeds; full search space and per-phase budgets archived with each run).

First, a \emph{baseline calibration} phase repeatedly runs the unsteeered pipeline over many seeds and records the mean, best, and standard deviation of the activation-side steering statistic produced by the code (the same separability gain we report as $\Delta\rho$ relative to a mean-difference baseline).
From this we derive a minimum gain threshold (a floor value together with a multiple of the baseline standard deviation) so that later stages ignore improvements indistinguishable from seed noise.

Second, a \emph{coarse search} draws a large batch of random configurations from a discrete grid over injection layer, PCA dimension, persistent-homology filtration knobs, component selection and weighting options, local negative-pool size, and subsampling controls; each configuration is evaluated by one end-to-end pipeline run, ranked by average gain, and only the top fraction is kept.

Third, a \emph{focused search} densifies the grid around those survivors by local moves that perturb several TDA-related coordinates (and occasionally the layer) to propose refined candidates, again ranked by the same activation objective.

Fourth, a \emph{robustness} phase re-evaluates the leading candidates on additional fresh seeds, aggregates per-candidate statistics, and discards settings whose mean gain does not exceed the calibrated baseline plus the minimum gain threshold.

Finally, a \emph{behavioral validation} phase generates baseline versus steered outputs on harmful prompts and scores them with external judges (e.g.\ Llama Guard and HarmBench classifiers, plus inexpensive substring refusal probes where desired); the \emph{final selection} then picks from the robust shortlist using this behavioral evidence (in our main experiment, ranking by the sum of Llama Guard and HarmBench success-rate deltas), freezing the resulting YAML configuration and steering vector for downstream evaluation.

\clearpage
\subsubsection{Layer sweeps plots}

\begin{figure*}[h]
    \centering
    \includegraphics[width=\textwidth]{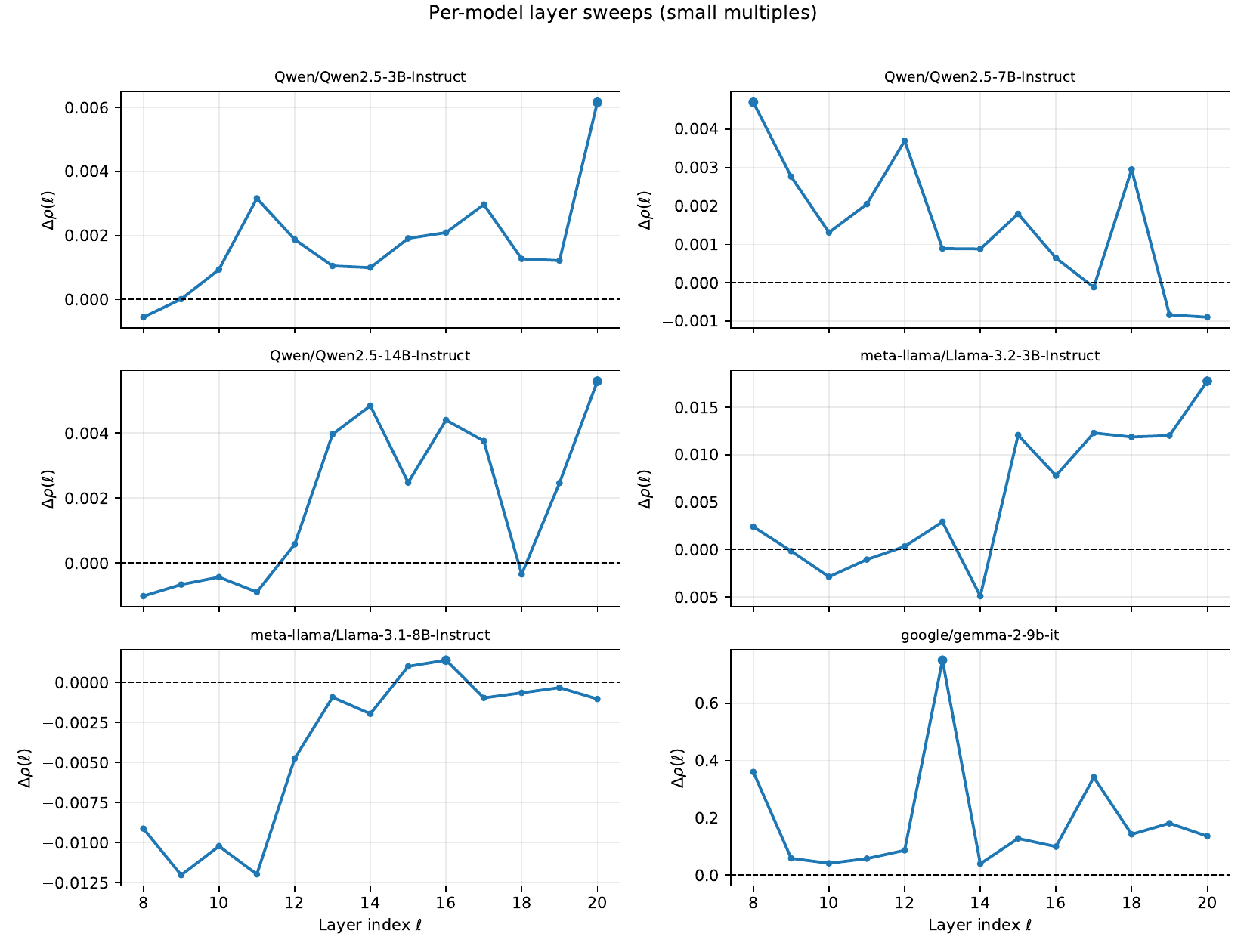}
    \caption{
    Per-model layer sweeps of $\Delta\rho(\ell)$ on the comparative run.
    Each panel reports one model, with the selected best layer marked.
    }
    \label{fig:appendix_layer_sweeps_subplots}
\end{figure*}

\begin{figure*}[h]
    \centering
    \includegraphics[width=\textwidth]{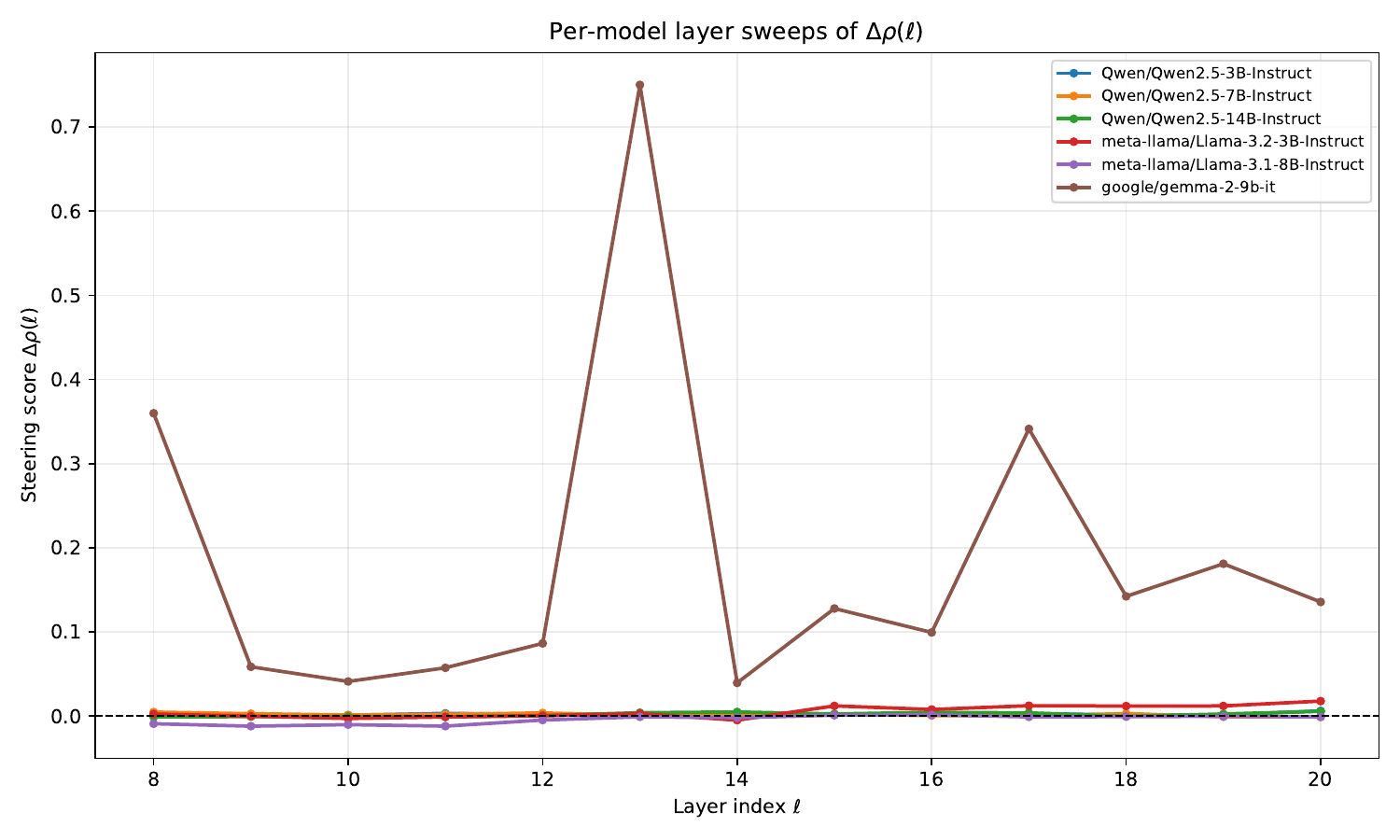}
    \caption{
    Overlay view of $\Delta\rho(\ell)$ across all models.
    }
    \label{fig:appendix_layer_sweeps_overlay}
\end{figure*}

\clearpage
\subsection{Statistical Significance}\label{subsec:ErrorPlots}

\begin{figure}[h]
  \centering
  \includegraphics[width=0.65\linewidth]{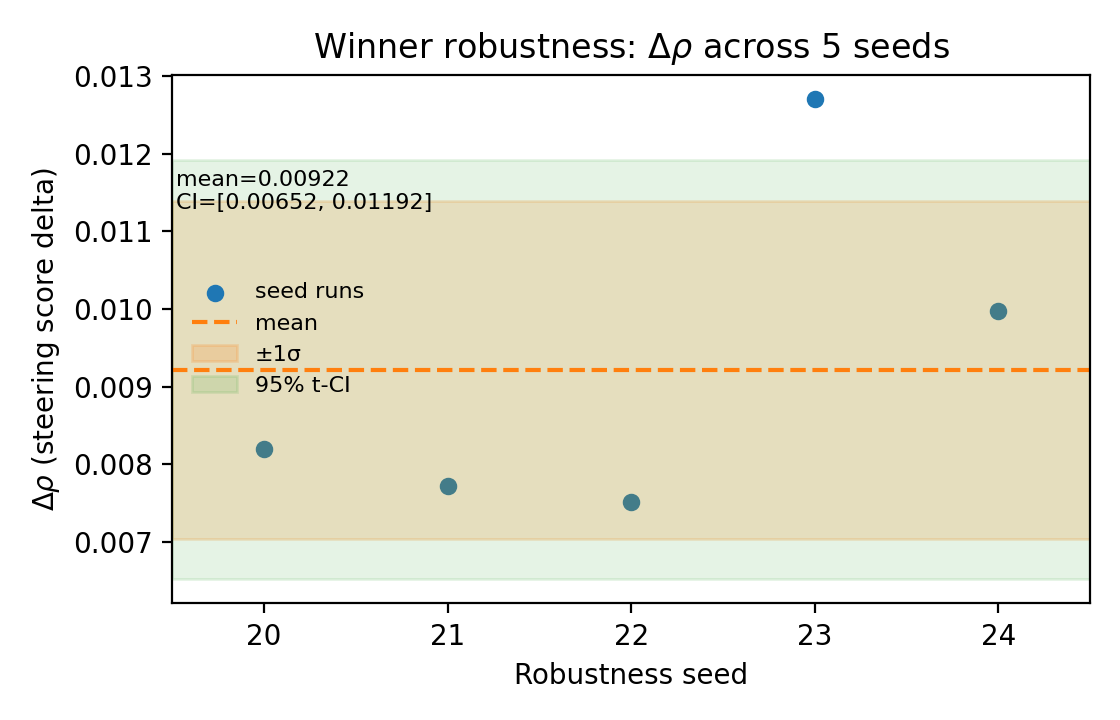}
  \caption{Robustness of the selected configuration across 5 random seeds. The shaded band shows $\pm 1\sigma$; the lighter band shows the 95\% $t$-confidence interval for the mean.}
  \label{fig:delta_rho_robustness_errorbars}
\end{figure}

Table~\ref{tab:delta_rho_robustness} and Figure~\ref{fig:delta_rho_robustness_errorbars} quantify how stable the observed steering gain is under randomness. In particular, $\Delta\rho$ denotes the steering-score delta (the improvement in the held-out refusal separation over the global mean-difference baseline) produced by the \emph{same} winning single-layer configuration, while only the random seed is changed. This isolates a key source of variability that can affect topological feature selection, subset extraction, and downstream direction construction, even when the data split, model architecture, and hyperparameters are held fixed.

Each table entry corresponds to one full re-run of the pipeline for the selected configuration. The five robustness seeds yield $\Delta\rho$ values of approximately $0.00752$--$0.01271$, with a mean of $\overline{\Delta\rho}\approx 0.00922$ and a sample standard deviation of $s\approx 0.00218$. The fact that \emph{all} five runs are positive indicates that the improvement is not driven by a single lucky run: the steering direction recovered from persistent-topology-guided selection consistently enhances linear separability between harmful and harmless activation clouds relative to the baseline. Moreover, the relative spread is modest; the standard deviation is about $24\%$ of the mean, suggesting that seed-induced fluctuations affect the magnitude of the gain but not its sign.

To further assess statistical significance, we report uncertainty around the mean. The table includes the standard error of the mean (SEM) and a 95\% $t$-confidence interval for the mean value of $\Delta\rho$ (with $n=5$ robustness runs). This interval is approximately $[0.00652,\ 0.01192]$, which remains clearly above zero. Under the standard $t$-interval assumptions (exchangeability of runs and reasonable approximation to symmetric errors), this supports the interpretation that the steering gain is robust and not merely an artifact of run-to-run noise.

Figure~\ref{fig:delta_rho_robustness_errorbars} provides an intuitive visual complement. The dashed horizontal line shows the mean across robustness seeds. The shaded darker band represents $\pm 1\sigma$ around the mean, illustrating the observed variability across seeds, while the lighter band shows the 95\% $t$-confidence interval for the mean. All seed points lie within the $\pm 1\sigma$ region, consistent with the numeric summary in Table~\ref{tab:delta_rho_robustness}. Overall, the combined evidence from the table and figure indicates that the persistent-topology-guided steering direction is stable: the method yields consistently positive improvements over the baseline across independent re-runs, and the uncertainty in the mean estimate does not overlap zero.

\begin{table}[h]
  \centering
  \small
  \caption{Robustness of the selected configuration across 5 random seeds in phase-3 robustness runs. $\Delta\rho$ is the steering-score improvement over the baseline direction.}
  \label{tab:delta_rho_robustness}
  \begin{tabular}{@{}l r@{}}
    \toprule
    Seed index (phase-3) & $\Delta\rho$ \\
    \midrule
    20 & 0.00819 \\
    21 & 0.00772 \\
    22 & 0.00752 \\
    23 & 0.01271 \\
    24 & 0.00997 \\
    \midrule
    Mean & 0.00922 \\
    Standard deviation (1$\sigma$) & 0.00218 \\
    Standard error of the mean & 0.00097 \\
    95\% $t$-CI (df $=4$) & $[0.00652,\ 0.01192]$ \\
    \bottomrule
  \end{tabular}
\end{table}

\clearpage
\subsection{Societal impact (positive and negative).}\label{subsec:SocialImpact}
This work studies inference-time steering methods intended to improve refusal behavior and reduce harmful responses in large language models, which can have positive societal impact by strengthening safety guardrails in high-risk interaction settings. In particular, improving controllable refusal may lower accidental assistance with dangerous requests and support safer deployment practices when combined with independent safety monitoring. At the same time, the same class of representation-level intervention techniques can be repurposed negatively: steering methods may be adapted to weaken existing safeguards, evade policy-aligned behavior, or induce model behaviors that are difficult to detect from surface text alone. Additional risks include over-reliance on automated judges, setup-dependent evaluation drift, and false confidence if activation-space gains are interpreted as universal safety guarantees. For these reasons, we frame our contribution as a diagnostic and methodological study under controlled benchmarks, not as a complete solution to safe deployment, and we encourage layered evaluation (multiple judges, robustness checks, and transparent reporting of failure modes) before real-world use.




\end{document}